\documentclass[twoside]{article} \usepackage{aistats2017}
\usepackage{color}
\usepackage{actionable} 
\usepackage{amsmath}
\usepackage{amssymb}
\usepackage{stmaryrd}
\usepackage{footmisc}
\usepackage{url}
\usepackage{tabularx}
\usepackage{subfigure}
\usepackage{graphicx}
\usepackage{epstopdf}
\usepackage{fancyhdr}
\usepackage{enumerate}
\usepackage{algpseudocode}
\usepackage{algorithm}
\usepackage{xspace}
\usepackage{mathtools, cuted}
\usepackage{tikz}
\usepackage{bm}
\usepackage{multirow,bigdelim}
\usepackage[colorlinks=true,urlcolor=blue]{hyperref}
\usepackage{pifont}
\usepackage{times}
\usepackage{booktabs}
\DeclareMathOperator*{\argmax}{arg\,max}
\usepackage{algpseudocode}
\newtheorem{theorem}{Theorem}

\newcommand{\xhdr}[1]{\vspace{1.7mm}\noindent{{\bf #1.}}}
\newcommand{\hide}[1]{}

\begin{document}

%

%

\twocolumn[

\aistatstitle{Learning Cost-Effective Treatment Regimes \\Using Markov Decision Processes}

\aistatsauthor{ Himabindu Lakkaraju \And Cynthia Rudin}

\aistatsaddress{ Stanford University \And Duke University } 
]

\begin{abstract}
Decision makers, such as doctors and judges, make crucial decisions such as recommending treatments to patients, and granting bails to defendants on a daily basis. Such decisions typically involve weighting the potential benefits of taking an action against the costs involved. 
In this work, we aim to automate this task of learning \emph{cost-effective, interpretable and actionable treatment regimes}. We formulate this as a problem of learning a decision list -- a sequence of if-then-else rules -- which maps characteristics of subjects (eg., diagnostic test results of patients) to treatments. We propose a novel objective to construct a decision list which maximizes outcomes for the population, and minimizes overall costs. We model the problem of learning such a list as a Markov Decision Process (MDP) and employ a variant of the Upper Confidence Bound for Trees (UCT) strategy which leverages customized checks for pruning the search space effectively.
Experimental results on real world observational data capturing judicial bail decisions and treatment recommendations for asthma patients demonstrate the effectiveness of our approach.

\end{abstract}

\section{Introduction}
\begin{figure}[ht!]
\centering
\scriptsize	
	\begin{tabular}{|l|}
		\hline \\
		\dsif \attrb{Spiro-Test}\dseq\val{Pos} \dsand \attr{Prev-Asthma}\dseq\val{Yes} \dsand \attr{Cough}\dseq\val{High} \dsthen \classh{C}  
 \\ \\
\dselif \attrb{Spiro-Test}\dseq\val{Pos} \dsand \attr{Prev-Asthma} \dseq \val{No} \dsthen \class{Q} \\ \\
\dselif \attr{Short-Breath} \dseq \val{Yes} \dsand \attr{Gender}\dseq \val{F} \dsand \attr{Age}\dsgeq \val{40} \dsand \attr{Prev-Asthma}\dseq \val{Yes} \dsthen \classh{C} \\ \\
\dselif \attrh{Peak-Flow}\dseq\val{Yes} \dsand \attr{Prev-RespIssue}\dseq\val{No} \dsand \attr{Wheezing} \dseq \val{Yes}, \dsthen \class{Q} \\ \\
\dselif \attr{Chest-Pain}\dseq\val{Yes} \dsand \attr{Prev-RespIssue} \dseq \val{Yes} \dsand \attra{Methacholine} \dseq \val{Pos} \dsthen \classh{C} \\ \\
\dselse \class{Q} \\ \\
		\hline
	\end{tabular}
	\caption{Regime for treatment recommendations for asthma patients output by our framework; \class{Q} refers to milder forms of treatment used for quick-relief, and \classh{C} corresponds to more intense treatments such as controller drugs (\classh{C} is higher cost than \class{Q}); Attributes in \attr{blue} are least expensive.}
	\label{fig:decisionlist}
    \vspace{-0.2in}
\end{figure}
Medical and judicial decisions can be complex: they involve careful assessment of the subject's condition, analyzing the costs associated with the possible actions, and the nature of the consequent outcomes. Further, there might be costs associated with the assessment of the subject's condition itself (e.g., physical pain endured during medical tests, monetary costs etc.). For instance, a doctor first diagnoses the patient's condition by studying the patient's medical history and ordering a set of relevant tests that are crucial to the diagnosis. In doing so, she also factors in the physical, mental and monetary costs incurred due to each of these tests.  Based on the test results, she carefully deliberates various treatment options, analyzes the potential side-effects as well as the effectiveness of each of these options. Analogously, a judge deciding if a defendant should be granted bail studies the criminal records of the defendant, and enquires for additional information (e.g., defendant's personal life or economic status) if needed. She then recommends a course of action that trades off the risk with granting bail to the defendant (the defendant may commit a new crime when out on bail) with the cost of denying bail (adverse effects on defendant, or defendant's family, cost of jail to the county). 

In practical situations, human decision makers often leverage personal experience to make decisions, without considering data, even if massive amounts of it exist for the problem at hand. There exist domains where machine learning models could potentially help -- but they would need to consider all three aspects discussed above: predictions of counterfactuals, costs of gathering information, and costs of treatments. Further, these models must be interpretable in order to create any reasonable chance of a human decision maker actually using them. In this work, we address the problem of learning such cost-effective, interpretable treatment regimes from observational data. 

Prior research addresses various aspects of the problem at hand in isolation. For instance, there exists a large body of literature on estimating treatment effects~\cite{d2007estimating,mcgough2009estimating,dorresteijn2011estimating}, recommending optimal treatments~\cite{abulesz1988novel,wallace2014personalizing,fan2016sequential}, and learning intelligible models for prediction~\cite{letham2015interpretable,lakkarajuinterpretable,lou2012intelligible,bien2009classification}. However, an effective solution for the problem at hand should ideally incorporate all of the aforementioned aspects. Furthermore, existing solutions for learning treatment regimes neither account for the costs associated with gathering the required information, nor the treatment costs. The goal of this work is to propose a framework which jointly addresses all of the aforementioned aspects.

We address the problem at hand by formulating it as a task of learning a decision list that maps subject characteristics to treatments (such as the one shown in Figure~\ref{fig:decisionlist}) such that it: 1) maximizes the expectation of a pre-specified outcome when used to assign treatments to a population of interest 2) minimizes costs associated with assessing subjects' conditions and 3) minimizes costs associated with the treatments themselves. We choose decision lists to express the treatment regimes because they are highly intelligible, and therefore, readily employable by decision makers. We propose a novel objective function to learn a decision list optimized with respect to the criterion highlighted above. We prove that the proposed objective is NP-hard by reducing it to the weighted exact cover problem. We then optimize this objective by modeling it as a Markov Decision Process (MDP) and employing a variant of the Upper Confidence Bound for Trees (UCT) strategy which leverages customized checks for pruning the search space effectively.

We empirically evaluate the proposed framework on two real world datasets: 1) judicial bail decisions 2) treatment recommendations for asthma patients. Our results demonstrate that the regimes output by our framework result in improved outcomes compared to state-of-the-art baselines at much lower costs. Further, the treatment regimes output by our approach are less complex and require fewer diagnostic checks to determine the optimal treatment. 

\section{Related Work}
Below, we provide an overview of related research on learning treatment regimes,  dynamic optimal treatment regimes, subgroup analysis, and interpretable models. \\

\textbf{Treatment Regimes.} The problem of learning treatment regimes has been extensively studied in the context of medicine and health care. Along the lines of \cite{BIOM:BIOM12354}, literature on treatment regimes can be categorized as: \emph{regression-based methods} and \emph{policy-search-based methods}. \textit{Regression-based methods}~\cite{qian2011performance,tian2014simple,robins1994correcting,vansteelandt2014structural,zhao2009reinforcement,qian2011performance,moodie2014q} model the conditional distribution of the outcomes given the treatment and characteristics of patients and choose the treatment resulting in the best possible outcome for each individual. \textit{Policy-search-based methods} search for a policy (a function which assigns treatments to individuals) within a pre-specified class of policies. The policy is chosen to optimize the expected outcome across the population of interest. Examples of such estimators include marginal structural mean models~\cite{robins1994correcting}, outcome weighted learning~\cite{zhao2012estimating,zhao2015doubly}, and robust marginal mean models~\cite{zhang2012robust,BIOM:BIOM12354}. Very few of the aforementioned solutions~\cite{BIOM:BIOM12354,moodie2012q} produce regimes which are intelligible. None of the aforementioned approaches explicitly account for treatment costs and costs associated with gathering information pertaining to patient characteristics.

While most work on learning treatment regimes has been done in the context of medicine, the same ideas apply to policies in other fields. 
To the best of our knowledge, this work is the first attempt in extending work on treatment regimes to judicial bail decisions. 

\xhdr{Dynamic Treatment Regimes} 
Recent research in personalized medicine has focused on developing \emph{dynamic treatment regimes}~\cite{laber2014dynamic,zhang2016interpretable,wallace2014personalizing,fan2016sequential}. The goal is to learn treatment regimes that maximize outcomes for patients in a given population by recommending a sequence of appropriate treatments over time, based on the state of the patient. There has been little attention paid to interpretability in this literature (with the exception of \cite{zhang2016interpretable}). None of the prior solutions for this problem consider treatment costs or costs associated with diagnosing a patient's condition. 

\xhdr{Subgroup Analysis} The goal of this line of research is to find out whether there exist subgroups of individuals in which a given treatment exhibits heterogeneous effects, and if so, how the treatment effect varies across them. This problem has been well studied~\cite{su2009subgroup,foster2011subgroup,loh2015regression,berger2014bayesian,imai2013estimating}. However, identifying subgroups with heterogeneous treatment effects does not readily provide us with  regimes. 

\xhdr{Interpretable Models} A large body of machine learning literature focused on developing interpretable models for classification~\cite{letham2015interpretable,lakkarajuinterpretable,lou2012intelligible,bien2009classification} and clustering~\cite{kim2014bayesian,lakkaraju2015bayesian,lakkaraju2016confusions}. To this end, various classes of models such as decision lists~\cite{letham2015interpretable}, decision sets~\cite{lakkarajuinterpretable}, prototype (case) based models~\cite{bien2009classification}, and generalized additive models~\cite{lou2012intelligible} were proposed. These classes of models were not conceived to model treatment effects. There has been recent work on leveraging decision lists to describe estimated treatment regimes~\cite{moodie2012q,laber2015tree,BIOM:BIOM12354}. These solutions do not account for the treatment costs or costs involved in gathering patient characteristics. They are also constructed using greedy methods, which causes issues with the quality of the models. 
\vspace{-0.1in}

\section{Our Framework}
First, we formalize the notion of treatment regimes and discuss how to represent them as decision lists. We then propose an objective function for constructing cost-effective treatment regimes. 

\subsection{Input Data and Cost Functions}

Consider a dataset $\mathcal{D} = \{(\textbf{x}_1, a_1, y_1),$  $(\textbf{x}_2, a_2, y_2)$ $\cdots$
$(\textbf{x}_N, a_N, y_N) \}$ comprised of $N$ independent and identically distributed observations, each of which corresponds to a \emph{subject} (individual), potentially from an observational study. 
Let $\textbf{x}_i = \left[ x^{(1)}_i, x^{(2)}_i, \cdots x^{(p)}_i \right] \in \left[\mathcal{V}_1, \mathcal{V}_2, \cdots \mathcal{V}_p\right]$ denote the \emph{characteristics} of subject $i$. $\mathcal{V}_f$ denotes the set of all possible values that can be assumed by a characteristic $f \in \mathcal{F} = \{1, 2, \cdots p\}$. Each characteristic $f \in \mathcal{F}$ can either be a binary, categorical or real valued variable.  In the medical setting, example characteristics include patient's age, BMI, gender, red blood cell count, glucose level etc., Let $a_i \in \mathcal{A} = \{1,2,\cdots m\}$ and $y_i \in \mathbbm{R}$ denote the \emph{treatment} assigned to subject $i$ and the corresponding \emph{outcome} respectively. We assume that $y_i$ is defined such that higher values indicate better outcomes. For example, the outcome of a patient can be regarded as a wellness improvement score that indicates the effectiveness of the assigned treatment. 

It can be much more expensive to determine certain subject characteristics compared to others. For instance, a patient's age can be easily retrieved either from previous records or by asking the patient. On the other hand, determining her glucose level requires more comprehensive testing, and is therefore more expensive in terms of monetary costs, time and effort required both from the patient as well as the clinicians. We assume access to a function $d: \mathcal{F} \rightarrow \mathbbm{R}$ which returns the cost of determining any characteristic in $\mathcal{F}$. The cost associated with a given characteristic $f \in \mathcal{F}$ is assumed to be the same for all the subjects in the population, though the framework can be extended to have patient-specific costs. Analogously, each treatment $a \in \mathcal{A}$ incurs a cost and we assume access to a function $d': \mathcal{A} \rightarrow \mathbbm{R}$ which returns the cost associated with treatment $a \in \mathcal{A}$.

We now discuss the notion of a treatment regime formally, and then introduce the class of models that we employ to express such regimes. 

\subsection{Treatment Regimes}
A treatment regime is a function which takes as input the characteristics of any given subject $\textbf{x}$ and maps them to an appropriate treatment $a 
\in \mathcal{A}$.
As discussed, prior studies~\cite{shiffman1997representation,marlowe2012adaptive} suggest that decision makers such as doctors and judges who make high stake decisions are more likely to trust, and, therefore employ models which are interpretable and transparent. We thus employ \emph{decision lists} to express treatment regimes (see example in Figure~\ref{fig:decisionlist}). A decision list is an ordered list of rules embedded within an if-then-else structure. A treatment regime\footnote{We use the terms decision list and treatment regimes interchangeably from here on.} expressed as a decision list $\pi$ 
\hide{takes the following form:\\
if $c_1$, then $a_1$ \\
else if $c_2$, then $a_2$ \\ 
$\cdot$ \\ 
$\cdot$ \\ 
else if $c_{L}$, then $a_{L}$\\
else $a_{L+1}$.

The above list } is a sequence of $L+1$ rules $\left[ r_1, r_2, \cdots, r_{L+1}\right]$. 
The last one, $r_{L+1}$, is a default rule which applies to all those subjects who do not satisfy any of the previous $L$ rules. Each rule $r_j$ (except the default rule) is a tuple of the form $(c_j, a_j)$ where $a_j \in \mathcal{A}$, and $c_j$ represents a \emph{pattern} which is a conjunction of one or more predicates. Each predicate takes the form $(f,o,v)$ where $f \in \mathcal{F}$, $o \in \{=, \neq, \leq, \geq, <, > \}$, and $v \in \mathcal{V}_f$ denotes some value $v$ that can be assumed by the characteristic $f$. For instance, ``Age $\geq$ 40 $\wedge$ Gender$=$Female" is an example of such a pattern. A subject $i$ is said to satisfy rule $j$ if his/her characteristics $x_i$ satisfy all the predicates in $c_j$. Let us formally denote this using an indicator function, $\textit{satisfy}(x_i, c_j)$ which returns a $1$ if $x_i$ satisfies $c_j$ and $0$ otherwise. 

The rules in $\pi$ partition the dataset $\mathcal{D}$ into $L + 1$ groups: $\{\mathcal{R}_1, \mathcal{R}_2 \cdots \mathcal{R}_{L}, \mathcal{R}_{\text{default}}\}$. A group $\mathcal{R}_j$, where $j \in \{1, 2, \cdots L\}$, is comprised of those subjects that satisfy $c_j$ but do not satisfy any of $c_1, c_2, \cdots c_{j-1}$. This can be formally written as:
\begin{equation}\label{eqn:rj} \mathcal{R}_j = \left\{ \textbf{x} \in \left[ \mathcal{V}_1 \cdots \mathcal{V}_p \right]\text{ } | \text{ satisfy}(\textbf{x}, c_j\} \wedge \bigwedge\limits_{t=1}^{j-1} \neg \text{ satisfy}(\textbf{x},c_t) \right\}.
\end{equation}

The treatment assigned to each subject by $\pi$ is determined by the group that he/she belongs to. For instance, if subject $i$ with characteristics $\mathbf{x}_i$ belongs to group $\mathcal{R}_j$ induced by $\pi$ i.e., $\mathbf{x}_i \in \mathcal{R}_j$, then subject $i$ will be assigned the corresponding treatment $a_j$ under the regime $\pi$. More formally,
\begin{equation}\label{eqn:pix}
\pi(\mathbf{x}_i) =  \sum\limits_{l=1}^{L} a_l \text{ } \mathbbm{1}(\textbf{x}_i \in \mathcal{R}_l) + a_\text{default} \text{ } \mathbbm{1}(\textbf{x}_i \in \mathcal{R}_\text{default}) 
\end{equation}
where $\mathbbm{1}$ denotes an indicator function that returns $1$ if the condition within the brackets evaluates to true and $0$ otherwise. Thus, $\pi$ returns the treatment assigned to $\mathbf{x}_i$.

Similarly, the cost incurred when we assign a treatment to the subject $i$ (\emph{treatment cost}) according to the regime $\pi$ is given by:
\begin{equation}\label{eqn:phi} \phi(\textbf{x}_i) = d'(\pi(\textbf{x}_i)) \end{equation}
where the function $d'$, defined in Section 3.1., takes as input a treatment $a \in \mathcal{A}$ and returns its cost.

We can also define the cost incurred in assessing the condition of a subject $i$ (\emph{assessment cost})  as per the regime $\pi$. Note that a subject $i$ belongs to the group $\mathcal{R}_j$ if and only if the subject does not satisfy the conditions $c_1 \cdots c_{j-1}$, but satisfies the condition $c_j$ (Refer to Eqn. \ref{eqn:rj}). To reach this conclusion, all the characteristics present in the corresponding antecedents $c_1 \cdots c_{j}$ must have been measured for subject $i$ and evaluated against the appropriate predicate conditions. This implies that the assessment cost incurred for this subject $i$ is the sum of the costs of all the characteristics that appear in $c_1 \cdots c_{j}$. If $\mathcal{N}_l$ denotes the set of all the characteristics that appear in $c_1 \cdots c_{l}$, the assessment cost of the subject $i$ as per the regime $\pi$ can be written as: 

\begin{equation}\label{eqn:psi}
\psi(\textbf{x}_i) = \sum\limits_{l=1}^{L} \left[ \mathbbm{1}(\textbf{x}_i \in \mathcal{R}_l) \times \left( \sum\limits_{e \in \mathcal{N}_l} d(e) \right) \right]. 
\end{equation}


\subsection{Objective Function}
We now formulate the objective function for learning a cost-effective 
treatment regime.
We first formalize the notions of expected outcome, assessment, and treatment costs of a treatment regime $\pi$ with respect to the dataset $\mathcal{D}$.

\paragraph{Expected Outcome} Recall that the treatment regime $\pi$ assigns a subject $i$ with characteristics $\textbf{x}_i$ to a treatment $\pi(\textbf{x}_i)$ (Equation \ref{eqn:pix}). The quality of the regime $\pi$ is partly determined by the expected outcome when all the subjects in $\mathcal{D}$ are assigned treatments according to $\pi$. The higher the value of such an expected outcome, the better the quality of the regime $\pi$. There is, however, one caveat to computing the value of this expected outcome -- we only observe the outcome $y_i$ resulting from assigning $\textbf{x}_i$ to $a_i$ in the data $\mathcal{D}$, and not any of the counterfactuals. If the regime $\pi$, on the other hand, assigns a different treatment $a' \neq a_i$ to $\mathbf{x}_i$, we cannot evaluate the policy on $\mathbf{x}_i$. 

The solutions proposed to compute expected outcomes in settings such as ours can be categorized as: adjustment by regression modeling, adjustment by inverse propensity score weighting, and doubly robust estimation. A detailed treatment of each of these approaches is presented in Lunceford et al.~\cite{lunceford2004stratification}. The success of regression based modeling and inverse weighting depends heavily on the postulated regression model and the postulated propensity score model respectively. In either case, if the postulated models are not identical to the true models, we have biased estimates of the expected outcome. On the other hand, doubly robust estimation combines the above approaches in such a way that the estimated value of the expected outcome is unbiased as long as one of the postulated models is identical to the true model. The doubly robust estimator for the expected outcome of the regime $\pi$, denoted by $g_1(\pi)$, can be written as: 
\begin{eqnarray}\label{eqn:g1}
g_1(\pi)= \frac{1}{N} \sum\limits_{i=1}^{N} \sum\limits_{a \in \mathcal{A}} o(i,a) \textrm{ where}
\end{eqnarray}
\begin{eqnarray*}
\lefteqn{o(i,a)=}\\
&&\left[ \frac{\mathbbm{1}(a_i = a)}{\hat{\omega}(x_i,a)} (y_i - \hat{y}(x_i, a)) +\hat{y}(x_i, a) \right]  
\mathbbm{1}(\pi(x_i) = a).
\end{eqnarray*}
\normalsize
\hide{
\begin{align}\label{eqn:g1}
g_1(\pi) = \frac{1}{N} \sum\limits_{i=1}^{N} \sum\limits_{a \in \mathcal{A}} \left[ \frac{\mathbbm{1}(a_i = a)}{\hat{\omega}(x_i,a)} \{y_i - \hat{y}(x_i, a)\} + \hat{y}(x_i, a) \right] \nonumber \\ 
\mathbbm{1}(\pi(x_i) = a)
\end{align}
}

$\hat{\omega}(x_i,a)$ denotes the probability that the subject $i$ with characteristics $x_i$ is assigned to treatment $a$ in the data $\mathcal{D}$. $\hat{\omega}$ represents the propensity score model. In practice, we fit a multinomial logistic regression model on $\mathcal{D}$ to learn this function. Our framework does not impose any constraints on the functional form of $\hat{\omega}$. Similarly, $\hat{y}(x_i,a)$ denotes the predicted outcome obtained as a result of assigning a subject characterized by $x_i$ to a treatment $a$. $\hat{y}$ corresponds to the outcome regression model and is learned in our experiments by fitting a linear regression model on $\mathcal{D}$ prior to optimizing for the treatment regimes. $\hat{y}$ and $\hat{\omega}$ could be modeled using any other method; this is an entirely separate step from the algorithm discussed here.

\paragraph{Expected Assessment Cost} 
 Recall that there are assessment costs associated with each subject. These costs are governed by the characteristics that will be used in assessing the subject's condition and recommending a treatment. 
The assessment cost of a subject $i$ treated using regime $\pi$ is given in Eqn.~\ref{eqn:psi}. The expected assessment cost across the entire population can be computed as: 
\begin{equation}\label{eqn:g2}
g_2(\pi) = \frac{1}{N} \sum\limits_{i=1}^{N} \psi(\mathbf{x}_i).
\end{equation}
It is important to ensure that our learning process favors regimes with smaller values of expected assessment cost. Keeping this cost low also ensures that the full decision list is sparse, which assists with interpretability.

\paragraph{Expected Treatment Cost} There is a cost associated with assigning treatment to any given subject. 
The treatment cost for a subject $i$ who is assigned treatment using regime $\pi$ is given in Eqn.~\ref{eqn:phi}. The expected treatment cost across the entire population can be computed as: 
\begin{equation}\label{eqn:g3}
g_3(\pi) = \frac{1}{N} \sum\limits_{i=1}^{N} \phi(\mathbf{x}_i).
\end{equation}
The smaller the expected treatment cost of a regime, the more desirable it is in practice. We present the complete objective function below.

\paragraph{Complete Objective}
We assume access to the following inputs: 1) the observational data $\mathcal{D}$; 2) a set $\mathcal{FP}$ of frequently occurring \emph{patterns} in $\mathcal{D}$. Recall that each pattern corresponds to a conjunction of one or more predicates. 
An example pattern is ``Age $\geq$ 40 $\wedge$ Gender$=$Female". In practice, such patterns can be obtained by running a frequent pattern mining algorithm such as Apriori~\cite{agrawal1994fast} on the set $\mathcal{D}$; 3) a set of all possible treatments $\mathcal{A}$. 

We define the set of all possible (pattern, treatment) tuples as $\mathcal{L} = \{ (c,a) | c \in \mathcal{FP}, a\in \mathcal{A}\}$ and $C(\mathcal{L})$ as the set of all possible combinations of $\mathcal{L}$. An element in $\mathcal{L}$ can be thought of as a rule in a decision list and an element in $C(\mathcal{L})$ can be thought of a list of rules in a decision list (without the default rule). We then search over all elements in the set $C(\mathcal{L}) \times \mathcal{A}$ to find a regime which maximizes the expected outcome (Eqn. \ref{eqn:g1}) while minimizing the expected assessment (Eqn. \ref{eqn:g2}), and treatment costs (Eqn. \ref{eqn:g3}) all of which are computed over $\mathcal{D}$. Our objective function can be formally written as: 
\begin{equation}\label{eqn:fullobj}
\argmax\limits_{\pi \in C(\mathcal{L}) \times \mathcal{A}} \lambda_1 g_1(\pi) - \lambda_2 g_2(\pi) - \lambda_3 g_3(\pi)
\end{equation}
where $g_1, g_2, g_3$ are defined in Eqns.~\ref{eqn:g1}, ~\ref{eqn:g2}, \ref{eqn:g3} respectively, and $\lambda_1$ and $\lambda_2$ are non-negative weights that scale the relative influence of the terms in the objective. 
\begin{theorem}
The objective function in Eqn. \ref{eqn:fullobj} is NP-hard. (Please see appendix for details.)
\end{theorem}
Note that NP-hardness is a worst case categorization only; with an efficient search procedure, it is practical to obtain a good approximation on most reasonably-sized datasets.
\subsection{Optimizing the Objective}
We optimize our objective by modeling it as as a Markov Decision Process (MDP) and then employing Upper Confidence Bound on Trees (UCT) algorithm to find a treatment regime which maximizes Eqn. \ref{eqn:fullobj}. We also propose and leverage customized checks for guiding the exploration of the UCT algorithm and pruning the search space effectively.

\paragraph{Markov Decision Process Formulation}
Our goal is to find a sequence of rules which maximize the objective function in Eqn. \ref{eqn:fullobj}.
To this end, we formulate a fully observable MDP such that the optimal policy of the posited formulation provides a solution to our objective function.

\begin{figure}
\centering
\includegraphics[scale=0.25]{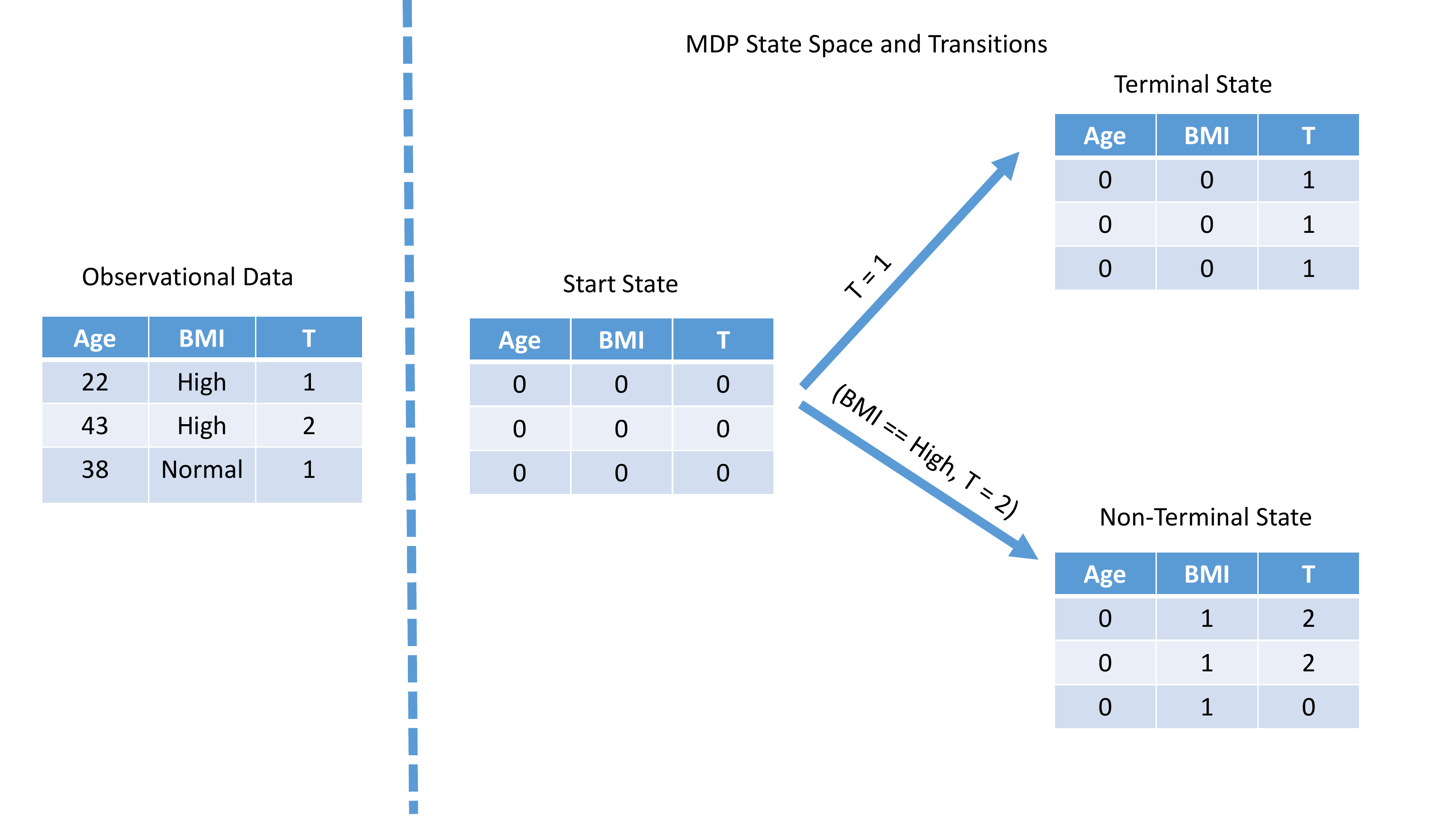}
\vspace{-0.30in}
\caption{Sample Observational Data and the corresponding Markov Decision Process Representation; T stands for Treatment.}
\label{fig:mdp}
\end{figure}

A fully observable MDP is characterized by a tuple $(\textbf{S}, \textbf{A}, \textbf{T}, \textbf{R})$ where $\textbf{S}$ denotes the set of all possible states, $\textbf{A}$ denotes the set of all possible actions, 
$\textbf{T}$ and $\textbf{R}$ represent the transition and reward functions respectively. Below we define each of these in the context of our problem. Figure \ref{fig:mdp} shows a snapshot of the state space and transitions for a small dataset. 

\emph{State Space.} Conceptually, each state in our state space captures the effect of some partial or fully constructed decision list. To illustrate, let us consider a partial decision list with just one rule ``if Age $\geq$ 40 $\wedge$ Gender $=$ Female, then T1". This partial list induces that: (i) all those subjects that satisfy the condition of the rule are assigned treatment T1, and (ii) Age and gender characteristics will be required in determining treatments for all the subjects in the population. 

To capture such information, we represent a state $\tilde{s} \in \boldsymbol{S}$ by a list of tuples $\left[ (\boldsymbol{\tau}_1(\tilde{s}), \sigma_1(\tilde{s})), \cdots (\boldsymbol{\tau}_N(\tilde{s}), \sigma_N(\tilde{s}))\right]$ where each tuple corresponds to a subject in $\mathcal{D}$.  $\boldsymbol{\tau}_i(\tilde{s})$ is a binary vector of length $p$ defined such that $\tau^{(j)}_i(\tilde{s}) = 1$ if the characteristic $j$ will be required for determining subject $i$'s treatment, and 0 otherwise. Further, $\sigma_i(\tilde{s})$ captures the treatment assigned to subject $i$. If no treatment has been assigned to $i$, then $\sigma_i(\tilde{s}) = 0$.

Note that we have a single start state $\tilde{s}_0$ which corresponds to an empty decision list. $\boldsymbol{\tau}_i(\tilde{s}_0)$ is a vector of $0$s, and $\sigma_i(\tilde{s}_0) = 0$ for all $i$ in $\mathcal{D}$ indicating that no treatments have been assigned to any subject, and no characteristics were deemed as requirements for assigning treatments. Furthermore, a state $\tilde{s}$ is regarded as a terminal state if for all $i$, $\sigma_i(\tilde{s})$ is non-zero indicating that treatments have been assigned to all the subjects.

\emph{Actions.} Each action can take one of the following forms: 1) a rule $r \in \mathcal{L}$, which is a tuple of the form (pattern, treatment). Eg., (Age$\geq$40 $\wedge$ Gender$=$Female, T1). This specifies that subjects who obey conditions in the pattern are prescribed the treatment. Such action leads to a non-terminal state. 2) a treatment $a \in \mathcal{A}$, which corresponds to  the default rule, thus this action leads to a terminal state. 

\emph{Transition and Reward Functions.} We have a deterministic transition function which ensures that taking an action $\tilde{a} = (\tilde{c}, \tilde{t}) $ from state $\tilde{s}$ will always lead to the same state $\tilde{s}'$. Let $U$ denote the set of all those subjects $i$ for which treatments have already been assigned to be in state $\tilde{s}$ i.e., $\sigma_i(\tilde{s}) \neq 0$ and let $U^{c}$ denote the set of all those subjects who have not been assigned treatment in the state $\tilde{s}$. Let $U'$ denote the set of all those subjects $i$ which do not belong to the set $U$ and which satisfy the condition $\tilde{c}$ of action $\tilde{a}$. Let $Q$ denote the set of all those characteristics in $\mathcal{F}$ which are present in the condition $\tilde{c}$ of action $\tilde{a}$. If action $\tilde{a}$ corresponds to a default rule, then $Q = \emptyset$ and $U'= U^{c}$. With this notation in place, the new state $\tilde{s'}$ can be characterized as follows: 1) $\tau^{(j)}_i(\tilde{s}') = \tau^{(j)}_i(\tilde{s})$ and $\sigma_i(\tilde{s}') = \sigma_i(\tilde{s})$ for all $i \in U$, $j \in \mathcal{F}$; 2) $\tau^{(j)}_i(\tilde{s}') = 1$ for all $i \in U^{c}$, $j \in Q$; 3) $\sigma_i(\tilde{s}') = \tilde{t}$ for all $i \in U'$.

Similarly, the immediate reward obtained when we reach $\tilde{s}'$ by taking $\tilde{a} = (\tilde{c}, \tilde{t})$ from the state $\tilde{s}$ can be written as: 
\[ 
\frac{\lambda_1}{N} \sum\limits_{i \in U'} o(i, \tilde{t}) - \frac{\lambda_2}{N}  \sum\limits_{i \in U^{c}, j \in Q} d(j) - \frac{\lambda_3}{N} \sum\limits_{i \in U'} d'(\tilde{t})
\]
where $o$ is defined in Eqn.~\ref{eqn:g1}, $d$ and $d'$ are cost functions for characteristics and treatments respectively (see Section 3.1).
\begin{table*}[t]
\centering
	\scriptsize
	\begin{tabular}{ c c c } 
     & \textbf{Bail Dataset} & \textbf{Asthma Dataset}\\
     \midrule
     \# of Data Points & 86152 & 60048\\
     \midrule
     Characteristics \& Costs & age, gender, previous offenses, prior arrests,  & age, gender, BMI, BP, short breath, temperature, \\
     & current charge, SSN (cost = 1) &  cough, chest pain, wheezing, past allergies, asthma history, \\
     & & family history, has insurance (cost 1)\\
     & marital status, kids, owns house, pays rent & peak flow test (cost = 2) \\
     & addresses in past years (cost = 2) & \\
     & & spirometry test (cost = 4) \\
     & mental illness, drug tests (cost = 6) & methacholine test (cost = 6) \\
     \midrule
     Treatments \& Costs & release on personal recognizance (cost = 20) & quick relief (cost = 10) \\
     & release on conditions/bond (cost = 40) & controller drugs (cost = 15) \\
     \midrule
     Outcomes \& Scores &  no risk (score = 100), failure to appear (score = 66) & no asthma attack for $\geq$ 4 months (score = 100) \\
     & non-violent crime (score = 33) & no asthma attack for 2 months (score = 66) \\
     & violent crime (score = 0) & no asthma attack for 1 month (score = 33) \\
     & & asthma attack in less than 2 weeks (score = 0) \\
	 \bottomrule
	\end{tabular}
\caption{Summary of datasets.}
\label{tab:datasets}
\vspace{-0.2in}
\end{table*}
\normalsize

\paragraph{UCT with Customized Pruning}
The basic idea behind the Upper Confidence Bound on Trees (UCT)~\cite{kocsis2006bandit} algorithm is to iteratively construct a search tree for some pre-determined number of iterations. At the end of this procedure, the best performing policy or sequence of actions is returned as the output. Each node in the search tree corresponds to a state in the MDP state space and the links in the tree correspond to the actions. UCT employs the UCB-1 metric~\cite{browne2012survey} for navigating through the search space. 

We employ a UCT-based algorithm for finding the optimal policy of our MDP formulation, though we leverage customized checks to further guide the exploration process and prune the search space. Recall that each non-terminal state in our state space corresponds to a partial decision list. We exploit the fact that we can upper-bound the value of the objective for any given partial decision list. The upper bound on the objective for any given non-terminal state $\tilde{s}$ can be computed by approximating the reward as follows: 1) all the subjects who have not been assigned treatments will get the best possible treatments without incurring any treatment cost 2) no additional assessments are required by any subject (and hence no additional assessment costs levied) in the population. The upper bound on the incremental reward is thus: 
\[\textrm{upper bound}(U^c)=\lambda_1 \frac{1}{N}\sum_{i \in U^{c}} \max_t o(i,t).
\]

During the execution of UCT procedure, whenever there is a choice to be made about which action needs to be taken, we employ checks based on the upper bound of the objective value of the resulting state. Consider a scenario in which the UCT procedure is currently in state $\tilde{s}$ and needs to choose an action. For each possible action $\tilde{a}$ (that does not correspond to a default rule\footnote{We can compute exact values of objective function if the action is a default rule because the corresponding decision list is fully constructed.}) from state $\tilde{s}$, we determine the upper bound on the objective value of the resulting state $\tilde{s}'$. If this value is less than either the highest value encountered previously for a complete rule list, or the objective value corresponding to the best default action from the state $\tilde{s}$, then we \emph{block} the action $\tilde{a}$ from the state $\tilde{s}$. This state is provably suboptimal.

\section{Experimental Evaluation}
Here, we discuss the detailed experimental evaluation of our framework. 
First we analyze the outcomes obtained and costs incurred 
when recommending treatments using our approach. 
Then, we present an ablation study which explores the contributions of each of the terms in our objective, followed by an analysis on real data.

\paragraph{Dataset Descriptions} 
Our first dataset consists of information pertaining to the 
\textbf{bail} decisions of about 86K defendants (see Table~\ref{tab:datasets}). It captures information about various defendant characteristics such as demographic attributes, past criminal history, personal and health related information for each of the 86K defendants. Further, the decisions made by judges in each of these cases (release without/with conditions) and the corresponding outcomes (e.g., if a defendant committed another crime when out on bail) are also available. 

We assigned costs to characteristics, and treatments based on discussions with subject matter experts. The characteristics that were harder to obtain were assigned higher costs compared to the ones that were readily available. Similarly, the treatment that placed a higher burden on both the defendant (release on condition) was assigned a higher cost. 
When assigning scores to outcomes, undesirable scenarios (e.g., violent crime when released on bail) received lower scores.

Our second dataset (Refer Table~\ref{tab:datasets}) captures details of about 60K \textbf{asthma} patients~\cite{lakkarajuinterpretable}. For each of these 60K patients, 
various attributes such as demographics, symptoms, past health history, test results have been recorded. Each patient in the dataset was prescribed either quick relief medications or long term controller drugs. Further, the outcomes in the form of time to the next asthma attack (after the treatment began) were recorded. The longer this interval, the better the outcome, and the higher the outcome score. 

We assigned costs to characteristics, and treatments based on the inconvenience (physical/mental/monetary) they caused to patients. 

\paragraph{Baselines}
We compared our framework to the following state-of-the-art treatment recommendation approaches: 1) Outcome Weighted Learning (OWL)~\cite{zhao2012estimating} 2) Modified Covariate Approach (MCA)~\cite{tian2014simple} 3) Interpretable and Parsimonious Treatment Regime Learning (IPTL)~\cite{BIOM:BIOM12354}. While none of these approaches explicitly account for treatment costs or costs required for gathering the subject characteristics, MCA and IPTL minimize the number of characteristics/covariates required for deciding the treatment of any given subject. OWL, on the other hand, utilizes all the characteristics available in the data when assigning treatments. 

\paragraph{Experimental Setting}
The objective function that we proposed in Eqn.~\ref{eqn:fullobj} has three parameters $\lambda_1, \lambda_2, \text{ and } \lambda_3$. These parameters could either be specified by an end-user or learned using a validation set. We set aside 5\% of each of our datasets as a validation set to estimate these parameters. We automatically searched the parameter space to find a set of parameters that produced a decision list with the maximum average outcome on the validation set (discussed in detail later) and satisfied some simple constraints such as: 1) average assessment cost $\leq$ 4 on both the datasets 2) average treatment cost $\leq$ 30 for the bail data; average treatment cost $\leq$ 12 for the asthma data. 
We then used a coordinate ascent strategy to search the parameter space and update each parameter $\lambda_j$ while holding the other two parameters constant. The values of each of these parameters were chosen via a binary search on the interval $(0, 1000)$. We ran the UCT procedure for our approach for 50K iterations. We used both gaussian and linear kernels for OWL and employed the tuning strategy discussed in Zhao et. al.~\cite{zhao2012estimating}. In case of IPTL, we set the parameter that limits the number of the rules in the treatment regime to $20$. We evaluated the performance of our model and other baselines using 10 fold cross validation.

\subsection{Quantitative Evaluation}
We analyzed the performance of our approach CITR (Cost-effective, Interpretable Treatment Regimes) on various aspects such as outcomes obtained, costs incurred, and intelligibility. We computed the following metrics:

\textbf{Avg. Outcome} Recall that a treatment regime assigns a treatment to every subject in the population. We used the prediction model $\hat{y}$ (defined in Section 3.3) to obtain an outcome score given the characteristics of the subject and the treatment assigned (we used ground truth outcome scores whenever they were available in the data). We computed the average outcome score of all the subjects in the population.\\
\textbf{Avg. Assess Cost} We determined assessment costs incurred by each subject based on what characteristics were used to determine their treatment. We then averaged all such per-subject assessment costs to obtain the average assessment cost. \\
\textbf{Avg. \# of Characs} We determined the number of characteristics that are used when assigning a treatment to each subject in the population and then computed the average of these numbers. \\
\textbf{Avg. Treat Cost} We computed the average of the treatment costs incurred by all the subjects in the population. \\
\textbf{List Len} Our approach CITR and the baseline IPTL express treatment regimes as decision lists. In order to compare the complexity of the resulting decision lists, we computed the number of rules in each of these lists.

While higher values of average outcome are preferred, lower values on all of the other metrics are desirable. 

\paragraph{Results} Table~\ref{tab:results} (top panel) presents the values of  the metrics computed for our approach as well as the baselines. It can be seen that the treatment regimes produced by our approach results in better average outcomes with lower costs across both datasets. 
While IPTL and MCA do not explicitly reduce costs, they do minimize the number of characteristics required for determining treatment of any given subject. Our approach produces regimes with the least \textit{cost} for a given average number of characteristics required to determine treatment (Avg. \# of Characs). It is also interesting that our approach produces more concise lists with fewer rules compared to the baselines. While the treatment costs of all the baselines are similar, there is some variation in the average assessment costs and the outcomes. IPTL turns out to be the best performing baseline in terms of the average outcome, average assessment costs, and average no. of characteristics. The last line of Table \ref{tab:results} shows the average outcomes and the average treatment costs computed empirically on the observational data. Both of our datasets are comprised of decisions made by human experts. It is interesting that the regimes learned by algorithmic approaches perform better than human experts on both of the datasets.   

\begin{table*}[t]
	\scriptsize
    \centering
	\begin{tabular}{ c c c c c c | c c c c c} 
     & \multicolumn{5}{c}{\textbf{Bail Dataset}} & \multicolumn{5}{c}{\textbf{Asthma Dataset}}\\
     \midrule
     & Avg. & Avg. & Avg. & Avg. \# of & List & Avg. & Avg. & Avg. & Avg. \# of & List \\
	 & Outcome & Assess Cost & Treat Cost & Characs. & Len & Outcome & Assess Cost & Treat Cost & Characs. & Len \\
     \midrule
     \textbf{CITR} & 79.2 & 8.88 & 31.09 & 6.38 & 7 & 74.38 & 13.87 & 11.81 & 7.23 & 6 \\
     IPTL & 77.6 & 14.53 & 35.23 & 8.57 & 9 & 71.88 & 18.58 & 11.83 & 7.87 & 8\\
     MCA & 73.4 & 19.03 & 35.48 & 12.03 & - & 70.32 & 19.53 & 12.01 & 10.23 & - \\ 
     OWL (Gaussian) & 72.9 & 28 & 35.18 & 13 & - & 71.02 & 25 & 12.38 & 16 & - \\
     OWL (Linear) & 71.3 & 28 & 34.23 & 13 & - & 71.02 & 25 & 12.38 & 16 & - \\
     \midrule
     CITR - No Treat & 80.5 & 8.93 & 34.48 & 7.57 & 7 & 77.39 & 14.02 & 12.87 & 7.38 & 7 \\
     CITR - No Assess & 81.3 & 13.83 & 32.02 & 9.86 & 10 & 78.32 & 18.28 & 12.02 & 8.97 & 9 \\
     CITR - Outcome & 81.7 & 13.98 & 34.49 & 10.38 & 10 & 79.37 & 18.28 & 12.88 & 9.21 & 9 \\   
     \midrule
     Human & 69.37 & - & 33.39 & - & - & 68.32 & - & 12.28 & - & -\\
	 \bottomrule
	\end{tabular}
\caption{Results for Treatment Regimes. Our approach: CITR; Baselines: IPTL, MCA, OWL; Ablations of our approach: CITR - No Treat, CITR - No Assess, CITR - Outcome; Human refers to the setting where judges and doctors assigned treatments. }
\label{tab:results}
\vspace{-0.15in}
\end{table*}
\normalsize

\subsubsection{Ablation Study}
We also analyzed the effect of various terms of our objective function on the outcomes, and the costs incurred. To this end, we experimented with three different ablations of our approach: 1) \emph{CITR - No Treat}, which is obtained by excluding the term corresponding to the expected treatment cost in our objective ($g_3(\pi)$ in Eqn. \ref{eqn:fullobj}). 2) \emph{CITR - No Assess}, which is obtained by excluding the expected assessment cost term in our objective ($g_2(\pi))$ in Eqn. \ref{eqn:fullobj}) 3) \emph{CITR - Outcome}, which is obtained by excluding both assessment and treatment cost terms from our objective. 

Table~\ref{tab:results} (second panel) shows the values of the metrics discussed earlier in this section for all the ablations of our model. Naturally, removing the treatment cost term increases the average treatment cost on both datasets. Naturally, removing the assessment cost part of the objective results in regimes with much higher assessment costs (8.88 vs$.$ 13.83 on bail data; 13.87 vs$.$ 18.28 on asthma data). The length of the list also increases for both the datasets when we exclude the assessment cost term. These results demonstrate that each term in our objective function is crucial to producing a cost-effective interpretable regime.


\subsection{Qualitative Analysis}
The treatment regimes produced by our approach on asthma and bail datasets are shown in Figures~\ref{fig:decisionlist} and~\ref{fig:decisionlist2} respectively. 

It can be seen in Figure~\ref{fig:decisionlist2} that methacholine test which is more expensive appears at the end of the regime. This ensures that only a small fraction of the population (8.23\%) is burdened by its cost. Furthermore, it turns out that though the spirometry test is slightly expensive compared to patient demographics and symptoms, it would be harder to determine the treatment for a patient without this test. This aligns with research on asthma treatment recommendations~\cite{pereira2015breath,boulet2015benefits}. Furthermore, it is interesting to note that the regime not only accounts for test results on spirometry and peak flow but also assesses if the patient has a previous history of asthma or respiratory issues. If the test results are positive and the patient has no previous history of asthma or respiratory disorders, then the patient is recommended quick relief drugs. On the other hand, if the test results are positive and the patient suffered previous asthma or respiratory issues, then controller drugs are recommended.

\begin{figure}[ht!]
\centering
\scriptsize	
	\begin{tabular}{|l|}
		\hline \\
		\dsif \attr{Gender}\dseq\val{F} \dsand \attr{Current-Charge} \dseq \val{Minor} \attr{Prev-Offense}\dseq\val{None} \dsthen \class{RP}  
 \\ \\
\dselif \attr{Prev-Offense}\dseq\val{Yes} \dsand \attr{Prior-Arrest} \dseq \val{Yes} \dsthen \classh{RC} \\ \\
\dselif \attr{Current-Charge} \dseq \val{Misdemeanor} \dsand \attr{Age}\dsleq \val{30} \dsthen \classh{RC} \\ \\
\dselif \attr{Age}\dsgeq\val{50} \dsand \attr{Prior-Arrest}\dseq\val{No}, \dsthen \class{RP} \\ \\
\dselif \attrh{Marital-Status}\dseq\val{Single} \dsand \attrh{Pays-Rent} \dseq \val{No} \dsand \attr{Current-Charge} \dseq \val{Misd.} \dsthen \classh{RC} \\ \\
\dselif \attrh{Addresses-Past-Yr}\dsgeq\val{5} \dsthen \classh{RC} \\ \\
\dselse \class{RP} \\ \\
		\hline
	\end{tabular}
	\caption{Treatment regime for bail data; \class{RP} refers to milder form of treatment: release on personal recognizance, and \classh{RC} is release on condition which is comparatively harsher.}
	\label{fig:decisionlist2}
    \vspace{-0.15in}
\end{figure}

In case of the bail dataset, the constructed regime is able to achieve good outcomes without even using the most expensive characteristics such as mental illness tests and drug tests.  Personal information characteristics, which are slightly more expensive than defendant demographics and prior criminal history, appear only towards the end of the list and these checks apply only to 21.23\% of the population. It is interesting that the regime uses the defendant's criminal history as well as personal and demographic information to make recommendations. For instance, females with minor current charges (such as driving offenses) and no prior criminal records are typically released on bail without conditions such as bonds or checking in with the police. On the other hand, defendants who have committed crimes earlier are only granted conditional bail. 

\section{Conclusions}
In this work, we proposed a framework for learning cost-effective, interpretable treatment regimes from observational data. To the best of our knowledge, this is the first solution to the problem at hand that addresses all of the following aspects: 1) maximizing the outcomes 2) minimizing the treatment costs, and costs associated with gathering information required to determine the treatment 3) expressing regimes using an interpretable model. We modeled the problem of learning a treatment regime as a MDP and employed a variant of UCT which prunes the search space using customized checks. We demonstrated the effectiveness of our framework on real world data from judiciary and health care domains. 

\bibliographystyle{plain}
\bibliography{actionable}

\section{Appendix}

\subsection{Proof for Theorem 1}
\textbf{Statement:} The objective defined in Eqn. 8 is NP-hard. \\ \\
\textbf{Proof:}
The rough idea behind this proof is to establish the connection between the objective in Eqn. (8) and weighted exact-cover problem. \\ \\

Our objective function is given by:
\begin{equation}\label{eqn:fullobj}
\argmax\limits_{\pi \in C(\mathcal{L}) \times \mathcal{A}} \lambda_1 g_1(\pi) - \lambda_2 g_2(\pi) - \lambda_3 g_3(\pi)
\end{equation}
The goal is to find a sequence of $(c,a)$ pairs where $c \in \{ \mathcal{FP} \cup \emptyset \}$ and $a \in \mathcal{A}$ which not only covers all the data points in the dataset but also maximizes the objective given above. Note that $c = \emptyset$ denotes a default rule. \\

$\mathcal{FP}$ represents a set of frequently occurring patterns each of which is a conjunction of one or more predicates.
Examples: \\
(1) Age $\geq$ 40 $\wedge$ Gender $=$ Female; \\
(2) BMI $=$ High; \\
(3) Gender $=$ M $\wedge$ BP $=$ High $\wedge$ Age $\leq$ 25 \\

Such patterns are provided as input to us. We have defined the set $\mathcal{L}$ as: $\mathcal{L} = \mathcal{FP} \times \mathcal{A}$. This implies that an element in the set $\mathcal{L}$ will be of the form: \\
(Age $\geq$ 40 $\wedge$ Gender $=$ Female, T1) i.e., each element in $\mathcal{L}$ is a rule. Our goal is now to find an ordered list of rules from $\mathcal{L}$ (let us ignore the default rule for a little while) which maximize the objective in Eqn.~\ref{eqn:fullobj}. \\

Let us assume the set $\mathcal{L}$ comprises of the following candidate rules: \\
(1) (Age $\geq$ 40 $\wedge$ Gender $=$ Female, T1)\\
(2) (Age $\geq$ 40 $\wedge$ Gender $=$ Female, T2) \\
(3) (BMI $=$ High, T1)\\
(4) (BMI $=$ High, T2)\\

Let us create a new set $\mathcal{L}'$ from $\mathcal{L}$ as follows: for each rule $(c,a)$ in $\mathcal{L}$, append the negations of conditions of all possible combinations of all the other rules in $\mathcal{L}$. Also include in the new set $\mathcal{L}'$, the set of all possible combinations of negations of conditions in all the rules in the set $\mathcal{L}$. Following our example above, the new set $\mathcal{L}'$ will look like this: \\
(1) (Age $\geq$ 40 $\wedge$ Gender $=$ Female, T1) \\
(2) (Age $\geq$ 40 $\wedge$ Gender $=$ Female, T2) \\
(3) ($\neg$(Age $\geq$ 40 $\wedge$ Gender $=$ Female), T1) \\
(4) ($\neg$(Age $\geq$ 40 $\wedge$ Gender $=$ Female), T2) \\
(5) ($\neg$ (BMI $=$ High) $\wedge$ Age $\geq$ 40 $\wedge$ Gender $=$ Female, T1) \\
(6) ($\neg$ (BMI $=$ High) $\wedge$ Age $\geq$ 40 $\wedge$ Gender $=$ Female, T2) \\
(7) (BMI $=$ High, T1)\\
(8) (BMI $=$ High, T2)\\
(9) ($\neg$(BMI $=$ High), T1)\\
(10) ($\neg$(BMI $=$ High), T2)\\
(11) ($\neg$ (Age $\geq$ 40 $\wedge$ Gender $=$ Female) $\wedge$ BMI $=$ High, T1) \\
(12) ($\neg$ (Age $\geq$ 40 $\wedge$ Gender $=$ Female) $\wedge$ BMI $=$ High, T2)\\
(13) ($\neg$(Age $\geq$ 40 $\wedge$ Gender $=$ Female) $\wedge$ $\neg$(BMI $=$ High), T1) \\
(14) ($\neg$(Age $\geq$ 40 $\wedge$ Gender $=$ Female) $\wedge$ $\neg$(BMI $=$ High), T2)\\

Now, the problem of finding an ordered sequence of rules on $\mathcal{L}$ (plus a default rule $a \in \mathcal{A}$) can now be posed as the problem of finding an unordered set of rules on $\mathcal{L}'$. To illustrate, let us consider a decision list constructed using $\mathcal{L}$ in the above example: \\

(1) (Age $\geq$ 40 $\wedge$ Gender $=$ Female, T1) \\
(2) T2 \\

This list can now be expressed as an unordered set using the elements in $\mathcal{L}'$ as follows: \\ 

(Age $\geq$ 40 $\wedge$ Gender $=$ Female, T1) \\
($\neg$(Age $\geq$ 40 $\wedge$ Gender $=$ Female), T2) \\

We have thus reduced the problem of finding an ordered list of rules to that of unordered set of rules on $\mathcal{L}'$. More specifically, the problem is now reduced to that of choosing a set of rules from the set $\mathcal{L}'$ such that 1) each data point/element in the data is covered exactly once 2) the objective function in Eqn.~\ref{eqn:fullobj} is maximized. This problem can be formally written as: \\

\begin{align}\label{eqn:cover}
\min_{j \in  \mathcal{L}'} \text{ } \Psi(j) \phi(j) \nonumber \\
\text{ s.t. } \sum\limits_{j: satisfy(x_i,c_j)} \phi(j) = 1 \text {   } \forall \textbf{  } i: (x_i, a_i, y_i) \in \mathcal{D} \nonumber \\
\phi(j) \in \{ 0, 1 \} \text{  } \forall j: r_j \in \mathcal{L}'
\end{align}

where $\phi(j)$ is an indicator function which is $1$ if the rule $r_j$ is chosen to be in the set cover. $\Psi(j)$ is the cost associated with choosing the rule $r_j = (c_j, a_j)$ which is defined as: 
\[ \Psi(j) = \sum\limits_{i: satisfy(x_i, c_j)} -\frac{\lambda_1}{N} o(i, a_j) + \frac{\lambda_2}{N} \sum\limits_{e \in c_j} d(e) + \frac{\lambda_3}{N} d'(a_j) \]

Note that we basically split our complete objective function across the rules that will be chosen to be part of the final set cover. Further, we are dealing with a minimization problem here, so we flip the signs of the terms in the objective (which is a maximization function). \\

Eqn.~\ref{eqn:cover} is the weighted exact cover problem. Since this problem is NP-Hard, our objective function is also NP-Hard.

\end{document}